\documentclass[sigconf,nonacm]{acmart}

\setcopyright{none}
\AtBeginDocument{%
  }

\usepackage{multirow}
\usepackage{enumitem}
\usepackage{float}
\usepackage{placeins}
\usepackage{tabularx}
\usepackage{colortbl}
\usepackage[most]{tcolorbox}
\definecolor{generalrow}{HTML}{E6ECFA}
\definecolor{financerow}{HTML}{E4F1DE}
\definecolor{bestblue}{HTML}{1F5CFF}
\definecolor{bestgreen}{HTML}{16802B}
\definecolor{promptblue}{HTML}{253B7A}
\definecolor{promptpink}{HTML}{D7658D}
\definecolor{exampleteal}{HTML}{2F6F73}
\definecolor{examplemuted}{HTML}{6A7480}
\newcolumntype{Y}{>{\raggedright\arraybackslash}X}

\newtcblisting{promptbox}[2]{%
  enhanced,
  breakable,
  colback=#2!6,
  colframe=#2!85!black,
  coltitle=white,
  colbacktitle=#2!85!black,
  title={#1},
  title after break={#1 (continued)},
  fonttitle=\bfseries,
  boxrule=0.9pt,
  arc=2pt,
  left=1mm,
  right=1mm,
  top=1mm,
  bottom=1mm,
  listing only,
  listing options={
    basicstyle=\ttfamily\scriptsize,
    breaklines=true,
    columns=fullflexible,
    keepspaces=true
  }
}

\newtcolorbox{labelbox}[2]{%
  enhanced,
  colback=#2!6,
  colframe=#2!85!black,
  coltitle=white,
  colbacktitle=#2!85!black,
  title={#1},
  fonttitle=\bfseries,
  boxrule=0.9pt,
  arc=2pt,
  left=1.2mm,
  right=1.2mm,
  top=1mm,
  bottom=1mm
}

\newtcolorbox{examplebox}[1]{%
  enhanced,
  breakable,
  colback=exampleteal!5,
  colframe=exampleteal!85!black,
  coltitle=white,
  colbacktitle=exampleteal!85!black,
  title={#1},
  fonttitle=\bfseries,
  boxrule=0.9pt,
  arc=2pt,
  left=1.4mm,
  right=1.4mm,
  top=1mm,
  bottom=1mm,
  before skip=0.7em,
  after skip=0.7em
}

\begin{document}

\title{From SEC Filings to Financial Signals: Evaluating LLMs on Stock Market Reactions and Analyst Forecasts}

\author{Shengkun Wang}
\affiliation{%
  \institution{Virginia Tech}
  \city{}
  \country{}}
\author{Taoran Ji}
\affiliation{%
  \institution{Texas A\&M University-Corpus Christi}
  \city{}
  \country{}}
\author{Linhan Wang}
\affiliation{%
  \institution{Virginia Tech}
  \city{}
  \country{}}
\author{Sattar Mansi}
\affiliation{%
  \institution{Virginia Tech}
  \city{}
  \country{}}
\author{Chang-Tien Lu}
\affiliation{%
  \institution{Virginia Tech}
  \city{}
  \country{}}
\author{Yinlin Chen}
\affiliation{%
  \institution{Virginia Tech}
  \city{}
  \country{}}
\renewcommand{\shortauthors}{Wang et al.}

\begin{abstract}
Form 10-Q MD\&A sections combine long-form management narrative with accounting and forward-looking information. Benchmarks for financial filings commonly evaluate question answering or numerical reasoning, leaving unclear whether outputs from large language models (LLMs) align with market and analyst outcomes. We evaluate prompted, zero-shot outputs from 13 general-purpose and finance-specialized LLMs on 45,964 Form 10-Q filing events from 2001--2024, linking MD\&A text to cross-sectional fundamental ranks, pre-filing analyst forecasts, and realized abnormal returns. The evaluation covers return ranking under five controlled information views and next-quarter EPS forecasting under direct and consensus-anchored interfaces. Relative to disclosure alone, adding cross-sectional fundamental ranks raises mean effective-date rank correlation (RankIC) by 0.0089 for effective-event-date returns and 0.0098 for following-day returns, whereas analyst-summary context provides no comparable average gain. Across the four disclosure-containing views, finance-specialized models show an effective-event-date group advantage but no statistically reliable following-day advantage, while general-purpose Qwen models produce the strongest individual return-ranking results. Under the full-information view, direct LLM forecasts average the 43.4th percentile of the analyst distribution and do not consistently beat individual analysts. Consensus anchoring raises the mean inside-range share from 41.1\% to 59.5\% but increases reported-EPS error.

\end{abstract}


\maketitle

\section{Introduction}
\label{sec:introduction}

\begin{figure}[t]
    \centering
    \includegraphics[width=\columnwidth]{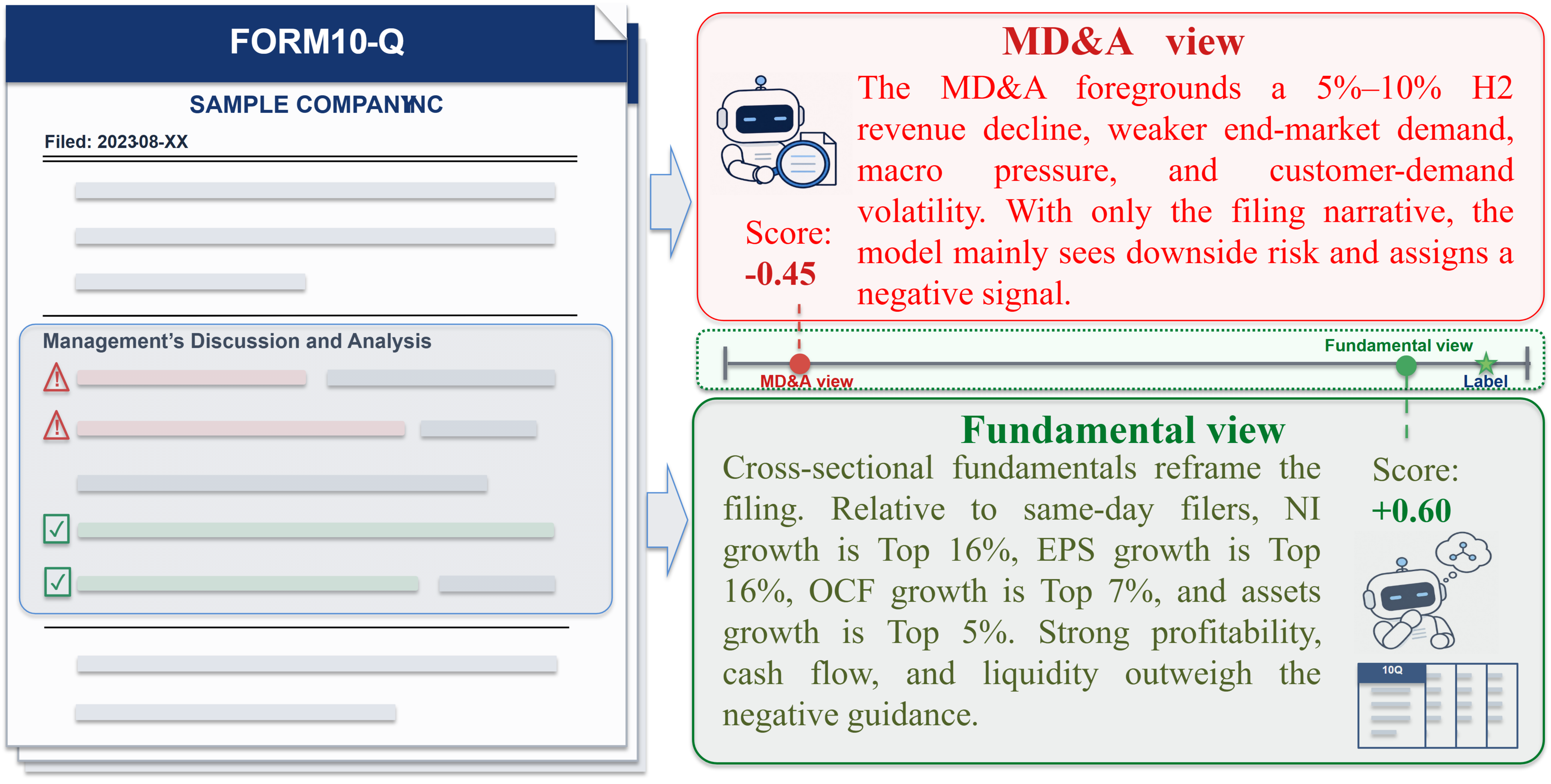}
    \caption{Adding cross-sectional fundamental ranks to the same 10-Q MD\&A can reverse the return signal inferred from the filing.}
    \Description{A schematic shows the same 10-Q MD\&A producing a negative return score under the disclosure-only view and a positive score after cross-sectional fundamental ranks are added.}
    \label{fig:context-dependent-filing-interpretation}
    \vspace{-0.9em}
\end{figure}

U.S. public firms disclose quarterly operating and financial information through Form 10-Q filings submitted to the SEC's EDGAR system \citep{sec_edgar}. 
Among these filings, the Management's Discussion and Analysis (MD\&A) section is particularly relevant for disclosure research because it combines management narrative, accounting quantities, risk discussion, and forward-looking statements \citep{li2010information}. 
Their language is also associated with future earnings and returns \citep{li2010information,cohen2020lazy}, providing observable outcomes outside the model.
This makes 10-Q MD\&A a natural test of whether language models can turn long-form disclosures into market-aligned signals; Figure~\ref{fig:context-dependent-filing-interpretation} shows how cross-sectional context can change the signal extracted from the same disclosure.

Existing financial NLP and LLM benchmarks mainly evaluate whether models can read, reason over, or answer questions about financial documents, while the conversion of such document understanding into market-aligned signals remains less studied. 
FinQA evaluates numerical reasoning over financial reports \citep{chen2021finqa}, and TAT-QA evaluates question answering over hybrid table-text financial content \citep{zhu2021tatqa}. 
SECQUE moves SEC filing evaluation toward expert-written financial analysis questions \citep{benyoash2025secque}. 
Fin-RATE further studies detail reasoning, cross-entity comparison, and longitudinal tracking over filings \citep{jiang2026finrate}. 
InvestorBench evaluates LLM-based agents in financial decision-making environments \citep{li2025investorbench}. 
These settings evaluate answer quality, reasoning behavior, or decisions in simulated financial environments. 
Our study instead asks how to measure whether models can turn 10-Q disclosures into financial signals.

This shift from document answers to financial signals creates a measurement problem. 
A useful target must be observable outside the model, tied to how financial users act on disclosures, and comparable across information sets and response formats. 
The problem is difficult in SEC-filing settings, where analysts combine filing text, accounting quantities, prior expectations, and firm context, while tasks may require cross-document, cross-entity, or longitudinal reasoning \citep{lai2025secqa,jiang2026finrate}. 
Model behavior may also vary with financial specialization \citep{wu2023bloomberggpt} and answer format \citep{zhao2021calibrate,pezeshkpour2024large}.

We turn this measurement problem into a controlled filing-to-signal evaluation with two financial prediction tasks.
First, the return-ranking task tests whether model-generated filing scores capture realized post-filing abnormal returns, a market-response target that also relates to asset-pricing evidence on firm characteristics and cross-sectional returns \citep{lewellen2015cross,gu2020empirical}.
Second, the analyst-forecasting task tests how LLM-generated EPS forecasts compare with professional analyst forecasts and reported EPS.

We provide comparative evidence on when and how signals generated by LLMs from long 10-Q filings align with market and analyst outcomes. Across matched filing events, three findings emerge. First, cross-sectional fundamental ranks provide the most consistent incremental gains in return alignment. Second, finance-specialized LLMs do not reliably dominate general-purpose models under the same information and evaluation setting. Third, LLM-generated EPS forecasts are consensus-centered and do not consistently beat individual analysts. Consensus anchoring moves forecasts closer to the professional forecast distribution while worsening accuracy against reported EPS. Our contributions are:

\begin{itemize}[leftmargin=*,itemsep=0pt,topsep=0.25em]
    \item We provide a controlled comparative filing-to-signal evaluation that links LLM outputs from long 10-Q
  disclosures to realized abnormal returns, professional analyst forecasts, and reported EPS.
    \item We show that adding cross-sectional fundamental ranks to identical MD\&A inputs
  significantly improves return-ranking alignment on both the effective event date and the following trading day,
  whereas pre-filing analyst expectations provide no comparable average gain.
    \item We compare general-purpose and finance-specialized LLMs under matched filing events, and show that financial-domain adaptation does not uniformly translate into stronger post-filing return alignment.
    \item We show that LLM-generated EPS forecasts remain close to consensus in aggregate but do
  not consistently outperform individual pre-filing analyst forecasts; consensus anchoring
  increases the share of forecasts within the analyst range while worsening accuracy against
  reported EPS.
\end{itemize}

\section{Related Work}

\subsection{Financial Text, Firm Characteristics, and Market Prediction}
Financial NLP and empirical asset pricing both study whether public information can be converted into predictive market signals. In asset pricing, firm characteristics are widely used to model the cross-section of expected returns, including characteristic-based forecasts and machine-learning models over stock-level predictors \citep{lewellen2015cross,gu2020empirical}. Disclosure-based studies show that SEC filings also contain market-relevant information: forward-looking MD\&A statements are associated with future earnings and returns \citep{li2010information}, and changes in filing language predict future returns and operations \citep{cohen2020lazy}.

Recent work also studies LLMs directly as return-prediction systems.  \citet{lopezlira2023chatgpt} derive short-horizon stock signals from news headlines, whereas \citet{wang2024stocktime} develop an LLM architecture for stock-price time series.  We connect this literature with disclosure-based return prediction by generating prompted, zero-shot scores from long 10-Q MD\&A under matched information views that add cross-sectional fundamental ranks or pre-filing analyst expectations.  We then evaluate cross-sectional alignment with abnormal returns among events sharing the same effective event date.  This design asks not only whether an LLM signal predicts returns, but how information-set construction affects measured return alignment.

\subsection{Financial LLMs and Analyst-Style Evaluation}
A separate line of work adapts language models to finance and evaluates financial reasoning. Domain-specific models such as FinBERT and BloombergGPT use financial corpora or financial-domain adaptation to improve financial signal processing \citep{liu2020finbert,wu2023bloomberggpt}, while FinGPT develops an open-source financial LLM framework \citep{yang2023fingpt}. Financial LLM benchmarks and datasets evaluate information extraction, numerical reasoning, question answering, forecasting, retrieval, and decision-oriented tasks over financial text, including FinQA, TAT-QA, PIXIU, FinBen, and open-book financial QA over public-company evidence in FinanceBench \citep{chen2021finqa,zhu2021tatqa,xie2023pixiu,xie2024finben,islam2023financebench}. SECQUE, SEC-QA, and Fin-RATE focus more directly on SEC-filing analysis \citep{benyoash2025secque,lai2025secqa,jiang2026finrate}. Work on LLM financial-statement analysis also compares model outputs with analyst forecasts using standardized financial statements \citep{kim2024financial}. Typical evaluation targets are answer correctness, retrieval, reasoning quality, or earnings-direction prediction. We evaluate model signals against realized abnormal returns, the professional analyst forecast distribution, and reported EPS.

\section{Task Formulation}
\label{sec:task-formulation}

Let $\mathcal{E}$ be a set of filing events, where each event $e=(i,d)$ corresponds to firm $i$ and original filing date $d$. The model observes an event-level input $x_e$ and returns a structured output that is parsed into a return-ranking score or an EPS forecast produced directly or as a revision to analyst consensus. Section~\ref{subsec:datasets} describes the filing samples and linked data sources used to construct $x_e$.

\subsection{Disclosure-to-Return Ranking}
\label{subsec:disclosure-to-return-ranking}
In the return-ranking task, let $t_e$ denote the \emph{effective event date} assigned from the filing date and SEC acceptance time as described below. Let $r_{i,t}$ denote firm $i$'s daily stock return on trading day $t$, and let $r^m_t$ denote the value-weighted market return on the same day. The primary target is the abnormal return on the effective event date,
\begin{equation}
    y_{e,0} = r_{i,t_e} - r^{m}_{t_e}.
\end{equation}
We refer to $y_{e,0}$ as the \emph{effective-event-date abnormal return}. We also define a following-trading-day companion target:
\begin{equation}
    y_{e,1} = r_{i,t_e+1} - r^{m}_{t_e+1}.
\end{equation}
Following prior studies of disclosure text and market outcomes \citep{li2010information,cohen2020lazy}, we use market-adjusted abnormal returns rather than raw returns. This choice removes broad market movements from the outcome and makes the target closer to the firm-specific response associated with each filing event.

The model produces a score $s_e$ for each filing event. Its cross-sectional alignment with abnormal returns is evaluated contemporaneously using $y_0$ and for persistence using following-day $y_1$. For two comparable events $e$ and $e'$ and target $q\in\{0,1\}$, the desired ordering is
\begin{equation}
    s_e > s_{e'} \quad \text{when} \quad y_{e,q} > y_{e',q}.
\end{equation}

\begin{figure}[t]
\centering
\includegraphics[width=\columnwidth]{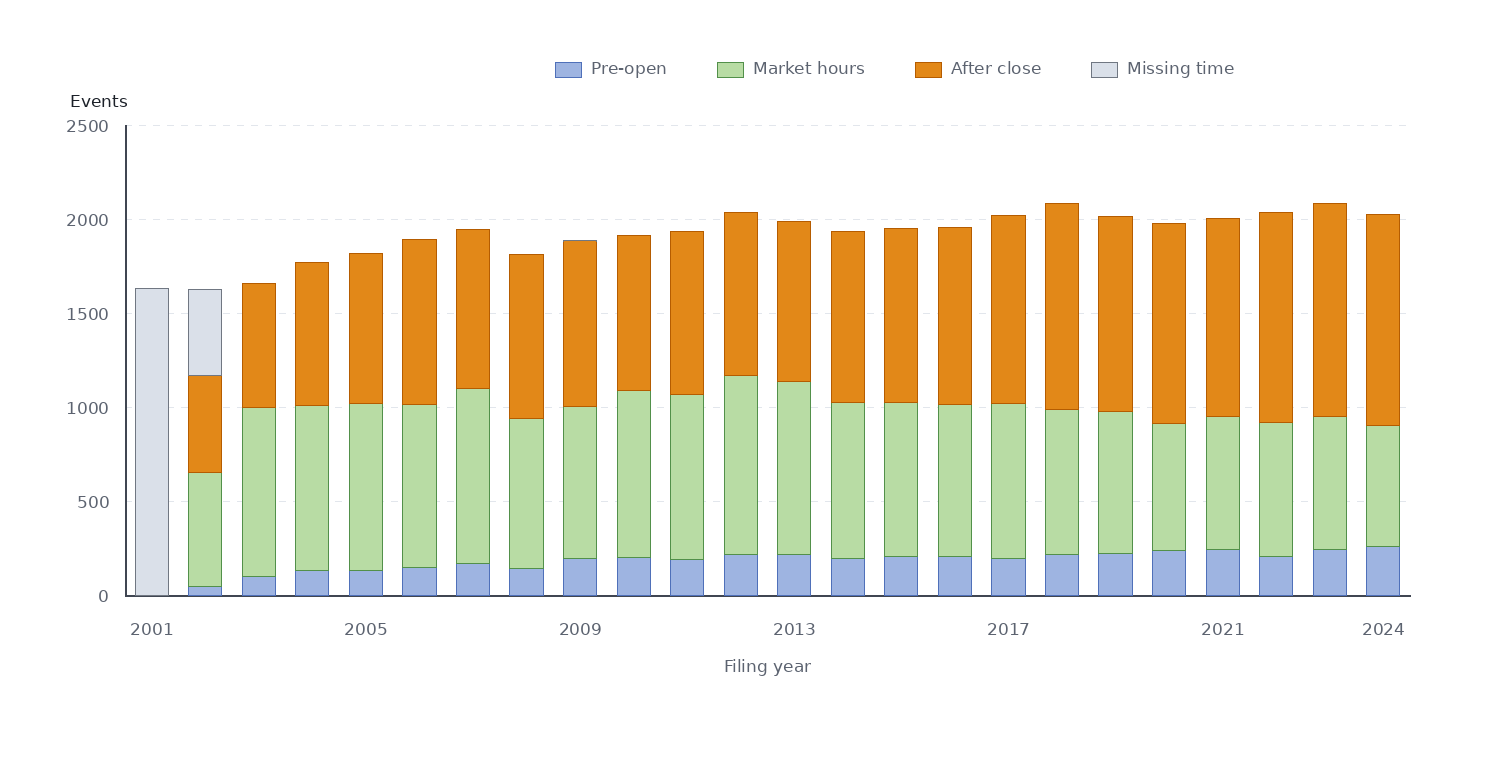}
\caption{Annual counts of 10-Q filing events by SEC acceptance-time bucket.}
\Description{A stacked bar chart shows annual 10-Q event counts from 2001 through 2024 divided into pre-open, market-hours, after-close, and missing-time filings. After-close filings form the largest category in most years.}
\label{fig:appendix-filing-time-mechanism}
\end{figure}

Figure~\ref{fig:appendix-filing-time-mechanism} shows that 10-Q filings can be accepted before the market opens, during regular trading hours, after the market closes, or without an observed acceptance time. Because the return targets use daily returns, the mapping from SEC filing time to the effective event date $t_e$ determines which daily return is assigned to each event. Let $d_e$ be the official filing date and let $\tau_e$ be the SEC acceptance timestamp when observed. If $d_e$ is a trading day and $\tau_e$ is observed before the regular market close, we set $t_e=d_e$; otherwise, we assign the event to the next trading day after $d_e$. Thus, every filing is public before the close of its effective-event-date session, and after-close disclosures are not matched to a session that ended before release.

\subsection{Analyst Forecasting}
\label{subsec:analyst-forecasting}

For analyst forecasting, the model predicts an earnings update from the filing. We use I/B/E/S Detail History forecasts (Institutional Brokers' Estimate System) for GPS earnings per share (EPS), where GPS is the I/B/E/S measure for generally accepted accounting principles (GAAP), or fully reported, EPS. For each event $e$, we select the first quarterly GPS EPS target period whose fiscal-period end follows the filing firm's fiscal-quarter end. Let $\mathcal{A}_e$ be the set of individual pre-filing analyst forecasts for the matched I/B/E/S ticker and target period, and let $c_e$ be their mean. The model observes the event-level input $x_e$ and produces either a direct GPS EPS forecast $\hat{a}_e$ or an adjustment $\Delta_e$ to the pre-filing consensus. For the consensus-anchored interface, the final forecast is
\begin{equation}
    \hat{a}_e = c_e + \Delta_e .
\end{equation}
Evaluation compares $\hat{a}_e$ with the pre-filing analyst distribution and matched I/B/E/S reported GPS EPS.

\section{Experiments}
\label{sec:experiments}
This section describes the data, model set, prompting inputs, and evaluation metrics used for the two tasks. We first define the 10-Q filing-event sample and linked financial data, then describe the evaluated LLMs, input settings, output formats, and outcome metrics.

\subsection{Setup}
\label{subsec:setup}

\subsubsection{Data.}
\label{subsec:datasets}
We construct filing-event data by aligning authoritative SEC disclosures \citep{sec_edgar} with financial databases accessed through Wharton Research Data Services \citep{wrds}.  The SEC 10-Q texts are publicly available and can be retrieved directly from EDGAR, while WRDS provides standardized access to licensed financial databases used for linking firm fundamentals, analyst expectations, and returns.  The design goal is to evaluate filing understanding in a market-response setting: each event connects what a firm disclosed, structured firm fundamentals, pre-filing analyst expectations, and the stock-market response observed afterward.  The main quarterly-report analysis uses a 10-Q return universe of 45,964 filing events from 2001--2024, built from Management's Discussion and Analysis (MD\&A) sections from SEC EDGAR and linked to Compustat firm fundamentals \citep{compustat}, I/B/E/S analyst expectations \citep{ibes}, and CRSP daily stock returns \citep{crsp}.  Market-adjusted abnormal returns use CRSP's daily value-weighted market index return including distributions, a market-capitalization-weighted return over CRSP-covered U.S. equities, as the market return.  The median 10-Q MD\&A span contains 6,494 words.

\subsubsection{Input Information Views.}
\label{subsec:input-views}

We construct controlled information views from three reusable blocks while holding the 10-Q filing events, response interface, and post-filing return targets fixed, so differences in return alignment reflect the information provided to the model.
The \textit{disclosure} block contains the 10-Q MD\&A text and represents the narrative information released by management. 
The \textit{fundamentals} block contains cross-sectional fundamental ranks computed among firms sharing the same original filing date.
The ranks cover current-quarter Compustat variables and their year-over-year changes for revenue, net income, EPS, assets, equity, debt, cash, operating cash flow, net margin, leverage, and cash-to-assets.
The block also reports cross-section size.
This block tests whether LLMs can ground qualitative management language in the firm's cross-sectional rank profile.
The \textit{analyst-expectations} block contains pre-filing I/B/E/S fields for the selected target period: estimate count, EPS consensus mean and median, dispersion, and recommendation statistics.  It provides a professional anchor available before the disclosure.
Combining these blocks yields the five evaluated views: disclosure only; fundamentals only; disclosure plus analyst expectations; disclosure plus fundamentals; and disclosure plus fundamentals plus analyst expectations.

\subsubsection{Models.}
\label{subsec:models}
We evaluate 13 LLMs in two descriptive groups:
\begin{itemize}[leftmargin=*,itemsep=0pt,topsep=0.25em]
    \item \textbf{General-purpose LLMs:}\par Qwen3-32B and Qwen3-8B \citep{qwen2025qwen3}; Qwen2.5-Instruct at 32B, 14B, and 7B \citep{qwen2024qwen25}; Llama-3.1-8B \citep{grattafiori2024llama3}; Mistral-NeMo-12B \citep{mistral2024nemo}; and the Llama-8B version of DeepSeek-R1-Distill \citep{deepseekai2025deepseekr1}.
    \item \textbf{Finance-specialized LLMs:}\par Fino1-14B and Fin-o1-8B \citep{qian2025fino1}, Fin-R1-7B \citep{liu2025finr1}, Llama-Fin-8B \citep{ke2025findap}, and ODA-Fin-RL-8B \citep{opendataarena2026odafin}.  These finance-adapted LLMs are built from general backbones: Qwen2.5-14B-Instruct for Fino1-14B, Qwen2.5-7B-Instruct for Fin-R1-7B, Meta-Llama-3-8B-Instruct for Llama-Fin-8B, and Qwen3-8B for Fin-o1-8B and ODA-Fin-RL-8B.
\end{itemize}

We select this panel along three dimensions relevant to model choice in financial workflows: domain specialization, model family, and parameter scale.
Qwen- and Llama-based models appear in both groups, enabling backbone- and scale-aligned comparisons of finance adaptation, while the broader coverage measures variation across model families and sizes from 7B to 32B.

\subsubsection{Prediction Interfaces.}
\label{subsec:prediction-interfaces}

\begin{figure}[t]
\centering
\includegraphics[width=\columnwidth]{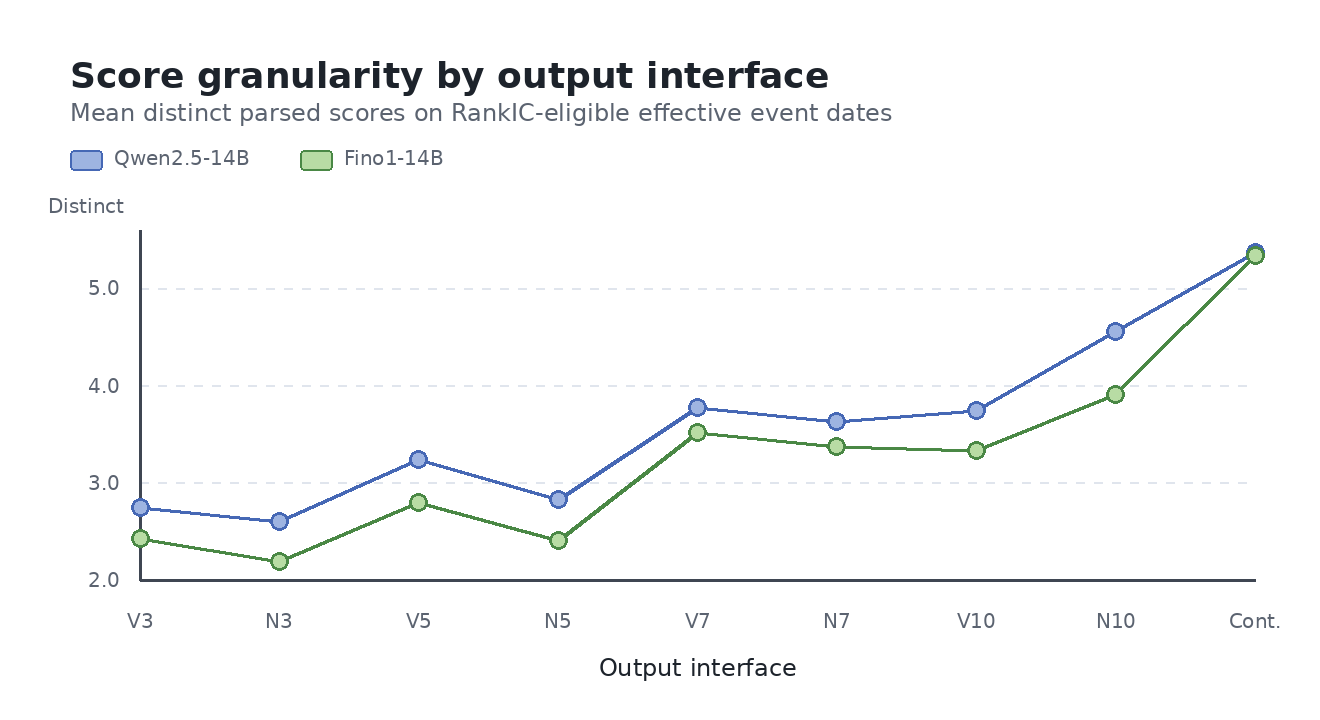}
\caption{Score granularity by response interface: mean distinct parsed scores on RankIC-eligible effective event dates. $V_k$ and $N_k$ denote $k$-level verbal and numeric ordinal interfaces; Cont. denotes continuous scoring.}
\Description{A line chart compares Qwen2.5-14B and Fino1-14B across verbal and numeric ordinal interfaces with 3, 5, 7, or 10 levels and a continuous interface. Distinct scores generally increase with interface resolution and are highest under continuous scoring.}
\label{fig:output-space-distinct}
\end{figure}

Output format is part of the measurement design because LLM behavior is sensitive to answer format \citep{zhao2021calibrate,pezeshkpour2024large}.  Return ranking uses $s_e\in[-1,1]$, where larger values indicate a more positive expected abnormal return; numeric and verbal ordinal formats are evaluated only in the interface ablation.  Figure~\ref{fig:output-space-distinct} shows that continuous scoring provides the finest within-date resolution, averaging about 5.3 distinct scores per eligible effective event date.  For GPS EPS forecasting, the direct interface outputs $\hat{a}_e$, while the consensus-anchored interface outputs $\Delta_e$ and sets $\hat{a}_e=c_e+\Delta_e$.

All interfaces use temperature 0 and require one prediction per event in the specified schema, with no additional prose.  Filing text is capped at 12,000 words using head--tail truncation that retains the final 2,000 words.  The parser reads JSON fields for return scores and consensus adjustments and a tagged \texttt{answer} block for direct forecasts; missing required fields invalidate an output.  Table~\ref{tab:prompt-output-schema} summarizes the three interfaces.

\subsubsection{Identity Masking and Context Matching.}
\label{subsec:identity-masking-design}
To distinguish identifier cues from event-level matching, we evaluate Qwen3-32B with masked $D$, masked $D$ plus matched $F$, and masked $D$ plus shuffled $F$.
Company names, tickers, dates, and four-digit years are replaced with typed placeholders; all other text and numbers remain.
Shuffling pools events by effective event date and original filing date and deranges whole $F$ blocks across different firms, preserving the RankIC and fundamental-rank cross-sections and each assigned rank profile.
Matched inputs are $(D_i,F_i)$; shuffled inputs are $(D_i,F_{\pi(i)})$ with $y_i$ fixed, breaking the event-level $D$--$F$ correspondence.

\subsubsection{Classical and supervised baselines.}
For classical baselines, we compute Loughran--McDonald net tone and its same-firm prior-quarter change \citep{loughran2016textual,cohen2020lazy}.  Over 2013--2024, Ridge predictors are fitted on expanding histories of realized return labels and tested on subsequent years.  Structured Ridge uses standardized, median-imputed fundamentals and analyst expectations; Text + Structured Ridge adds MD\&A features.  Qwen3-32B receives the corresponding full view without label fitting.  Headline RankICs use each method's eligible effective dates; paired LLM--Ridge contrasts use their shared dates.

\subsection{Evaluation}
\label{subsec:evaluation-protocol}

Return ranking is evaluated within effective event dates; EPS forecasts are compared with pre-filing analyst forecasts and reported GPS EPS.

\subsubsection{Return-ranking metric.}
\label{subsec:return-evaluation}
For target $q\in\{0,1\}$ and effective event date $t$, let $\mathcal{E}_{t,q}$ contain events with $t_e=t$ and valid $s_e$ and $y_{e,q}$.  We retain dates with at least five events, two distinct scores, and two distinct returns; $\mathcal{D}_q$ denotes this date set.

The effective-date RankIC is the Spearman correlation between score and return ranks within an effective event date:
\begin{equation}
    \rho_{t,q}
    =
    \operatorname{corr}\left(
        \operatorname{rank}_{e\in\mathcal{E}_{t,q}}(s_e),
        \operatorname{rank}_{e\in\mathcal{E}_{t,q}}(y_{e,q})
    \right),
    \label{eq:date-rankic}
\end{equation}
We report its average across effective event dates,
\begin{equation}
    \operatorname{RankIC}_q = \frac{1}{|\mathcal{D}_q|}
    \sum_{t\in\mathcal{D}_q}\rho_{t,q},
    \label{eq:mean-rankic}
\end{equation}
and its information ratio,
\begin{equation}
    \operatorname{RankICIR}_q
    =
    \frac{\operatorname{RankIC}_q}
    {\operatorname{sd}_{t\in\mathcal{D}_q}(\rho_{t,q})}.
    \label{eq:rankicir}
\end{equation}
RankIC is standard for cross-sectional return-ranking evaluation \citep{yang2020qlib}.  We complement it with a top-minus-bottom score-quintile return spread that measures return magnitude, following the portfolio-sort convention in \citet{gu2020empirical}.  Within each effective event date, we sort events into five equal-count score quintiles and compute
\begin{equation}
    \operatorname{LS}_q = \frac{1}{|\mathcal{D}_q|}
    \sum_{t\in\mathcal{D}_q}\left(\bar{y}^{+}_{t,q}-\bar{y}^{-}_{t,q}\right),
\end{equation}
where $\bar{y}^{+}_{t,q}$ and $\bar{y}^{-}_{t,q}$ are the highest- and lowest-score-quintile means.  All return-ranking inference treats effective event dates, rather than individual filing events, as the statistical observations.  One-sample $t$-statistics use the effective-date RankIC series $\{\rho_{t,q}\}$.  Information-view contrasts average paired within-model RankIC differences across models within each effective event date.  Model-group contrasts are the finance-specialized group mean minus the general-purpose group mean within each view and effective event date; reported contrast means and paired $t$-statistics use these date-level differences.

\subsubsection{Analyst metrics.}
\label{subsec:forecast-evaluation}
For each forecast-evaluable event, $\hat{a}_e$ is the parsed model forecast and $g_e$ is the matched I/B/E/S reported GPS EPS.  We report mean absolute error $|\hat{a}_e-g_e|$, position within the pre-filing analyst distribution, and analyst-relative beat rates.

To locate each model forecast within the contemporaneous analyst distribution, we compute an analyst-equivalent percentile:
\begin{equation}
    p_e
    =
    100\cdot
    \frac{1}{|\mathcal{A}_e|}
    \sum_{b\in\mathcal{A}_e}\mathbf{1}\{b\leq \hat{a}_e\},
\end{equation}
Lower percentiles indicate more pessimistic forecasts.  We also report whether $\hat{a}_e$ lies below, inside, or above the analyst range.  Analyst-relative beat rates compare model absolute error with the same-event median, 25th-percentile, and minimum individual-analyst errors \citep{kim2024financial}.

\begin{table*}[t]
\centering
\caption{Main 10-Q return-ranking results. IC is mean effective-date RankIC, and ICIR is its information ratio. $y_0$ is the effective-event-date abnormal return; $y_1$ is the following-trading-day abnormal return. Column maxima are blue (general-purpose) or green (finance-specialized); runners-up are underlined, and boldface marks the metric-wide maximum.}
\label{tab:information-view-full-results}
\tiny
\setlength{\tabcolsep}{1.35pt}
\renewcommand{\arraystretch}{1.06}
\resizebox{\textwidth}{!}{%
\begin{tabular}{l|rrrr|rrrr|rrrr|rrrr|rrrr}
\toprule
\textbf{Information View} & \multicolumn{4}{c|}{Disclosure only} & \multicolumn{4}{c|}{Fundamentals only} & \multicolumn{4}{c|}{Disclosure + expectations} & \multicolumn{4}{c|}{Disclosure + fundamentals} & \multicolumn{4}{c}{\scalebox{0.86}{Disclosure + fundamentals + expectations}} \\
\cmidrule(lr){1-1}\cmidrule(lr){2-5}\cmidrule(lr){6-9}\cmidrule(lr){10-13}\cmidrule(lr){14-17}\cmidrule(lr){18-21}
\textbf{Model} & $\mathrm{IC}_{y_0}$ & $\mathrm{IC}_{y_1}$ & \scalebox{0.82}{$\mathrm{ICIR}_{y_0}$} & $\mathrm{LS}_{0}$ & $\mathrm{IC}_{y_0}$ & $\mathrm{IC}_{y_1}$ & \scalebox{0.82}{$\mathrm{ICIR}_{y_0}$} & $\mathrm{LS}_{0}$ & $\mathrm{IC}_{y_0}$ & $\mathrm{IC}_{y_1}$ & \scalebox{0.82}{$\mathrm{ICIR}_{y_0}$} & $\mathrm{LS}_{0}$ & $\mathrm{IC}_{y_0}$ & $\mathrm{IC}_{y_1}$ & \scalebox{0.82}{$\mathrm{ICIR}_{y_0}$} & $\mathrm{LS}_{0}$ & $\mathrm{IC}_{y_0}$ & $\mathrm{IC}_{y_1}$ & \scalebox{0.82}{$\mathrm{ICIR}_{y_0}$} & $\mathrm{LS}_{0}$ \\
\midrule
\rowcolor{generalrow}
\multicolumn{21}{c}{General-purpose LLMs} \\
Qwen3-32B & \textcolor{bestblue}{0.07068} & 0.02640 & \textcolor{bestblue}{0.24043} & \textcolor{bestblue}{0.00707} & 0.06019 & 0.03025 & 0.19626 & \underline{0.00613} & 0.06124 & \underline{0.03200} & 0.19077 & 0.00634 & \underline{0.07555} & \underline{0.03906} & \textbf{\textcolor{bestblue}{0.25582}} & \underline{0.00761} & \textcolor{bestblue}{0.07447} & 0.03426 & \textcolor{bestblue}{0.23678} & 0.00564 \\
Qwen3-8B & 0.04513 & \textcolor{bestblue}{0.04402} & 0.13885 & 0.00533 & 0.05507 & 0.02764 & 0.18312 & 0.00388 & 0.05198 & \textcolor{bestblue}{0.03972} & 0.16100 & 0.00601 & 0.06279 & \textbf{\textcolor{bestblue}{0.05742}} & 0.19573 & 0.00705 & 0.06146 & \textcolor{bestblue}{0.05179} & 0.18890 & 0.00670 \\
Qwen2.5-32B & 0.05557 & \underline{0.02875} & 0.18114 & 0.00573 & 0.05914 & 0.02995 & 0.20532 & 0.00538 & 0.06204 & 0.02648 & 0.19842 & 0.00665 & 0.06384 & 0.03217 & 0.20596 & 0.00668 & 0.06331 & \underline{0.03507} & 0.20453 & 0.00645 \\
Qwen2.5-14B & \underline{0.05935} & 0.02048 & \underline{0.19028} & \underline{0.00677} & \textcolor{bestblue}{0.06618} & \underline{0.03652} & \textcolor{bestblue}{0.21219} & 0.00394 & \textcolor{bestblue}{0.06604} & 0.02355 & \textcolor{bestblue}{0.21016} & \underline{0.00696} & \textbf{\textcolor{bestblue}{0.07668}} & 0.03449 & 0.24074 & \textbf{\textcolor{bestblue}{0.00762}} & 0.06859 & 0.03144 & 0.21608 & \textcolor{bestblue}{0.00716} \\
Qwen2.5-7B & 0.04587 & 0.01679 & 0.14451 & 0.00520 & 0.06132 & 0.03527 & 0.19063 & \textcolor{bestblue}{0.00657} & 0.03804 & 0.01637 & 0.11883 & 0.00433 & 0.05607 & 0.02790 & 0.17612 & 0.00621 & 0.05226 & 0.02140 & 0.16355 & 0.00542 \\
Llama-3.1-8B & 0.04624 & 0.00709 & 0.14876 & 0.00443 & 0.04697 & 0.02400 & 0.14456 & 0.00441 & 0.03542 & 0.01480 & 0.11254 & 0.00419 & 0.03826 & 0.01330 & 0.12373 & 0.00399 & 0.04533 & 0.01961 & 0.14197 & 0.00370 \\
Mistral-NeMo-12B & 0.02539 & -0.00505 & 0.08251 & 0.00357 & 0.04044 & 0.01684 & 0.12998 & 0.00236 & 0.02031 & -0.00759 & 0.06732 & 0.00074 & 0.02476 & -0.00622 & 0.07800 & 0.00125 & 0.02538 & -0.00483 & 0.08344 & 0.00159 \\
DeepSeek-R1-Distill-Llama-8B & 0.03299 & 0.01080 & 0.11820 & 0.00212 & 0.05341 & 0.03295 & 0.18601 & 0.00529 & 0.02090 & 0.00257 & 0.07588 & -0.00127 & 0.02562 & 0.01118 & 0.09041 & -0.00117 & 0.03679 & 0.01221 & 0.12766 & -0.00057 \\
\midrule
\rowcolor{financerow}
\multicolumn{21}{c}{Finance-specialized LLMs} \\
Fino1-14B & 0.05697 & 0.01938 & 0.17904 & 0.00633 & 0.06194 & 0.03181 & 0.19777 & 0.00320 & 0.06141 & 0.02326 & 0.19819 & 0.00633 & 0.06790 & 0.03016 & 0.21130 & 0.00651 & 0.06524 & 0.03439 & 0.20503 & 0.00648 \\
Fin-R1-7B & 0.05713 & 0.02198 & 0.18360 & 0.00564 & 0.04098 & 0.02265 & 0.13037 & 0.00152 & 0.04127 & 0.02215 & 0.13131 & 0.00443 & 0.05751 & 0.02942 & 0.17959 & 0.00681 & 0.05615 & 0.03455 & 0.18521 & 0.00248 \\
Llama-Fin-8B & 0.04670 & 0.00589 & 0.14642 & 0.00516 & 0.04324 & 0.00480 & 0.13963 & 0.00219 & 0.05368 & 0.01837 & 0.17048 & 0.00506 & 0.05587 & 0.01033 & 0.17546 & 0.00435 & 0.05586 & 0.00843 & 0.17335 & 0.00336 \\
Fin-o1-8B & 0.05490 & 0.02211 & 0.17072 & 0.00580 & \underline{0.06397} & \textcolor{bestgreen}{0.04276} & \underline{0.21153} & 0.00461 & 0.05989 & 0.02056 & 0.18747 & 0.00683 & 0.07032 & 0.03263 & 0.22323 & 0.00717 & 0.07004 & 0.03389 & 0.22294 & \underline{0.00684} \\
ODA-Fin-RL-8B & 0.05901 & 0.02317 & 0.18535 & 0.00653 & 0.06138 & 0.03436 & 0.19400 & 0.00422 & \underline{0.06600} & 0.02661 & \underline{0.20719} & \textcolor{bestgreen}{0.00711} & 0.07153 & 0.03828 & \underline{0.24966} & 0.00722 & \underline{0.07199} & 0.03299 & \underline{0.22725} & 0.00638 \\
\bottomrule
\end{tabular}}
\end{table*}

\section{Results}
\label{sec:results}

We report three findings on information sets, model specialization, and analyst-relative forecasting.

\subsection{Cross-Sectional Context Improves Return Alignment}
\label{subsec:information-view-results}

Table~\ref{tab:information-view-full-results} reports the main 10-Q return-ranking results across five controlled information views while holding the filing events, response interface, and return targets fixed.  This design isolates whether disclosure text, cross-sectional fundamental ranks, and pre-filing analyst expectations contribute distinct return-ranking information.

Both unstructured disclosure and cross-sectional fundamental ranks contain standalone signal.  All 13 models produce positive $y_0$ RankIC from disclosure alone.  The fundamentals-only view produces positive $y_0$ and $y_1$ RankIC for all 13 models.  Qwen2.5-14B achieves the highest $y_0$ RankIC of 0.06618, whereas Fin-o1-8B achieves the highest $y_1$ RankIC of 0.04276.

The paired effective-date comparisons isolate the source of the incremental gain.  Disclosure plus fundamentals improves $y_0$ RankIC for 10 of 13 models and $y_1$ RankIC for 12 of 13 models relative to disclosure alone.  Adding cross-sectional fundamental ranks increases $y_0$ RankIC by 0.0089 ($t=2.55$, $p=0.011$) and $y_1$ RankIC by 0.0098 ($t=2.84$, $p=0.005$).  Under this combined view, Qwen2.5-14B attains the highest $y_0$ RankIC of 0.07668, Qwen3-8B the highest $y_1$ RankIC of 0.05742, and Qwen3-32B the highest $y_0$ RankICIR of 0.25582.  Cross-sectional fundamental ranks add return-ranking information beyond the filing text alone.

The gain is specific to the type of context rather than to adding more prompt content.  Relative to disclosure alone, adding analyst expectations changes $y_0$ RankIC by only $-0.0003$ ($t=-0.11$, $p=0.914$) and $y_1$ RankIC by 0.0003 ($t=0.09$, $p=0.924$).  Adding expectations on top of disclosure and fundamentals changes $y_0$ RankIC by $-0.0014$ ($t=-0.61$, $p=0.542$) and $y_1$ RankIC by $-0.0050$ ($t=-2.12$, $p=0.034$).  Across these comparisons, the reliable incremental gain comes from cross-sectional fundamental ranks.  Analyst expectations do not improve average return alignment.

\begin{figure*}[t]
\centering
\includegraphics[width=\textwidth]{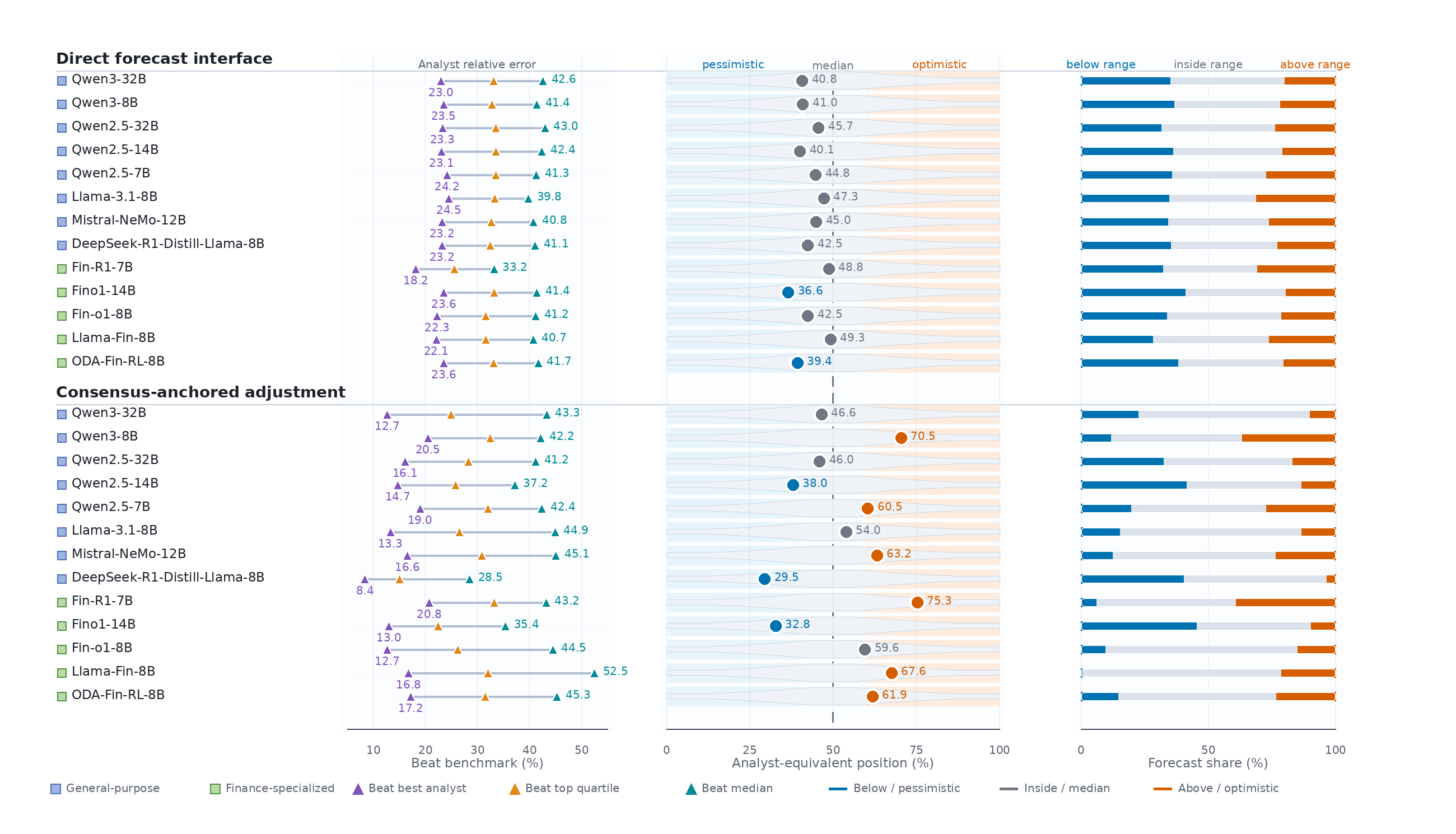}
\caption{Analyst-relative forecast position and reported-EPS accuracy under direct and consensus-anchored interfaces. Left: shares beating the median, top-quartile, and best individual analyst; center: analyst-equivalent percentile; right: shares below, inside, and above the pre-filing analyst range.}
\Description{Two model-by-model panels compare direct forecasts with consensus-anchored adjustments. The left panel reports analyst-relative beat rates, the center reports analyst-equivalent percentiles, and the right reports shares below, inside, and above the analyst range. Anchoring generally places more forecasts inside the analyst range but lowers the rate of beating the best analyst.}
\label{fig:analyst-distribution-position}
\end{figure*}

\paragraph{Masked and shuffled results.}
Table~\ref{tab:identity-calendar-masking} reports the conditions defined in Section~\ref{subsec:identity-masking-design} and includes the original disclosure-plus-fundamentals view as an unmasked reference.

\begin{table}[t]
\centering
\caption{Qwen3-32B RankIC under original and identity-masked input configurations. D denotes disclosure and F denotes cross-sectional fundamental ranks. Input-configuration $t$-statistics test mean effective-date RankIC against zero; the paired contrast reports matched-minus-shuffled $\Delta$RankIC and its $t$-statistic.}
\label{tab:identity-calendar-masking}
\small
\setlength{\tabcolsep}{3.3pt}
\renewcommand{\arraystretch}{1.08}
\begin{tabularx}{\columnwidth}{@{}Xrrrr@{}}
\toprule
& \multicolumn{2}{c}{$y_0$} & \multicolumn{2}{c}{$y_1$} \\
\cmidrule(lr){2-3}\cmidrule(l){4-5}
\textbf{Input configuration} & \textbf{RankIC} & \boldmath{$t$} & \textbf{RankIC} & \boldmath{$t$} \\
\midrule
Original D + F & 0.07555 & 9.319 & 0.03906 & 4.963 \\
\midrule
Masked D & 0.06163 & 7.852 & 0.02073 & 2.583 \\
Masked D + matched F & 0.07170 & 8.618 & 0.02776 & 3.596 \\
Masked D + shuffled F & 0.04935 & 6.104 & 0.01147 & 1.453 \\
\midrule
\textbf{Paired contrast} & \boldmath{$\Delta$}\textbf{RankIC} & \textbf{Paired }\boldmath{$t$} & \boldmath{$\Delta$}\textbf{RankIC} & \textbf{Paired }\boldmath{$t$} \\
Matched F $-$ shuffled F & 0.02235 & 2.937 & 0.01629 & 2.446 \\
\bottomrule
\end{tabularx}
\end{table}

Among the masked inputs, matched fundamentals produce the strongest RankIC: 0.07170 ($t=8.618$) on $y_0$ and 0.02776 ($t=3.596$) on $y_1$, versus 0.06163 and 0.02073 for disclosure alone and 0.04935 and 0.01147 with shuffled fundamentals. The matched-minus-shuffled $\Delta$IC is 0.02235 on $y_0$ (paired $t=2.937$) and 0.01629 on $y_1$ (paired $t=2.446$), making the difference statistically reliable at both horizons. Significant return alignment persists after company names, tickers, dates, and years are masked, indicating that the signal does not depend on recalling firm-specific outcomes from explicit identifiers. The matched-over-shuffled advantage further ties the incremental signal to the correct event-level correspondence between disclosure and fundamentals.

\paragraph{Baseline comparisons.}
Classical baselines test whether dictionary tone explains the signal. Loughran--McDonald net tone measures the positive-minus-negative word share, while net-tone change subtracts the same firm's prior-quarter value. Their $y_0$ RankICs are 0.0192 and 0.0263, respectively, well below the LLM result.

The supervised comparison is stricter because the Ridge models learn from realized return labels over expanding historical windows. On the common 2013--2024 test period, Qwen3-32B reaches a $y_0$ RankIC of 0.1009, versus 0.0714 for Structured Ridge and 0.0735 for Text + Structured Ridge. On shared effective dates, its paired advantages are 0.0340 over Structured Ridge ($t=3.13$, $p=0.002$) and 0.0319 over Text + Structured Ridge ($p=0.003$). The prompted, zero-shot LLM signal therefore outperforms both label-trained Ridge variants on the primary effective-event-date target.

The comparison also concerns the output available to users. In our implementation, Ridge yields a scalar prediction, whereas Qwen3-32B returns each score with short, event-specific textual evidence referring to the supplied filing and structured context. Human analysts can compare that evidence with the source information and assess whether the stated rationale is economically plausible. The resulting interface couples the return-ranking score with evidence for human review.

\subsection{Finance-Specialized LLMs Do Not Reliably Dominate General-Purpose Models}
\label{subsec:finance-specialization-return-results}

\paragraph{Group-level effects depend on the return horizon.}
Paired effective-date contrasts compare finance-specialized and general-purpose models under matched information views, with positive $\Delta\mathrm{IC}$ favoring the finance-specialized group. Across the four disclosure-containing views, this group leads on $y_0$ by 0.0062 under disclosure only ($t=2.08$, $p=0.038$), 0.0116 under disclosure plus expectations ($t=3.81$, $p<0.001$), 0.0129 under disclosure plus fundamentals ($t=4.26$, $p<0.001$), and 0.0096 under the full-information view ($t=3.21$, $p=0.001$). On $y_1$, the corresponding differences are $-0.0001$ ($t=-0.04$, $p=0.970$), 0.0041 ($t=1.32$, $p=0.188$), 0.0021 ($t=0.70$, $p=0.483$), and 0.0039 ($t=1.30$, $p=0.195$); none is statistically reliable. Financial specialization therefore provides a group advantage on $y_0$, but not on $y_1$.

\paragraph{Backbone- and family-aligned gains remain model dependent.}
Fino1-14B trails Qwen2.5-14B in all five $y_0$ views and exceeds it in only one of five $y_1$ views. Fin-o1-8B and ODA-Fin-RL-8B exceed Qwen3-8B in all five $y_0$ views but trail it in four of five $y_1$ views. Fin-R1-7B exceeds Qwen2.5-7B in four of five views at both horizons; Llama-Fin-8B exceeds Llama-3.1-8B in four of five $y_0$ views but only one $y_1$ view. General-purpose Qwen models also attain every Table~\ref{tab:information-view-full-results} maximum: $y_0$ RankIC of 0.07668, $y_1$ RankIC of 0.05742, $y_0$ RankICIR of 0.25582, and $\mathrm{LS}_0$ of 0.00762.

\paragraph{Scale and horizon effects remain mixed.}
Across the five views, Qwen3-32B beats Qwen3-8B on $y_0$ but trails it in four $y_1$ views; Qwen2.5-14B beats Qwen2.5-32B on $y_0$. As a longer-horizon robustness check under the full-information view, we compute RankIC using cumulative abnormal returns over trading days $t_e+1$ through $t_e+5$. The five leading models are Qwen3-32B, ODA-Fin-RL-8B, Fin-o1-8B, Qwen2.5-7B, and Fino1-14B, with RankICs of 0.03561, 0.03064, 0.02782, 0.02471, and 0.02237, respectively. Neither specialization nor scale dominates across horizons.

\subsection{LLM Forecasts Are Consensus-Centered but Not Consistently Analyst-Beating}
\label{subsec:analyst-forecasting-results}

Figure~\ref{fig:analyst-distribution-position} compares alignment with the pre-filing analyst distribution and accuracy against reported GPS EPS.

\paragraph{Direct forecasts cluster near the analyst center.}
Under the full view, the analyst-equivalent percentile averages 43.4 across models and ranges from 36.6 to 49.3, placing direct forecasts near but slightly below the analyst median. Yet only 41.1\% fall inside the analyst range, versus 35.0\% below and 23.9\% above. Mean percentile measures aggregate directional centering, whereas inside-range coverage measures event-level agreement. Direct forecasts exhibit the former but not consistently the latter.

\paragraph{Consensus anchoring improves distributional alignment but worsens reported-EPS error.}
Anchoring raises the inside-range share from 41.1\% to 59.5\%, or 18.432 percentage points ($t=6.08$, $p<0.001$), and the mean analyst-equivalent percentile from 43.4 to 54.3, or 10.886 points ($t=3.04$, $p=0.010$). Model-level percentiles nevertheless widen from 36.6--49.3 to 29.5--75.3, and mean absolute error against reported EPS rises by 10.798 ($t=9.41$, $p<0.001$). The wider percentile range shows that a common anchor does not homogenize model behavior; models differ in how they translate filing evidence into $\Delta_e$. Combined with the higher error, the increase in inside-range coverage reveals a shrinkage trade-off. Forecasts become more analyst-like at the cost of realized-earnings accuracy.

\paragraph{LLMs do not consistently beat individual analysts.}
The strongest direct models beat the median individual analyst on roughly 40--43\% of events with reported EPS and the best analyst on roughly 22--25\%. With anchoring, only Llama-Fin-8B exceeds a 50\% beat-median rate, at 52.5\%. Across models, anchoring raises the beat-median rate by only 1.163 percentage points ($t=0.64$, $p=0.535$) and lowers the beat-best rate by 7.390 points ($t=-6.16$, $p<0.001$). LLM-generated EPS forecasts resemble professional forecasts without consistently outperforming them.

\FloatBarrier

\section{Conclusion}
\label{sec:conclusion}

We evaluate 13 prompted, zero-shot LLMs across controlled information views and prediction interfaces, linking long 10-Q MD\&A to abnormal returns, analyst forecasts, and reported EPS. Adding cross-sectional fundamental ranks to disclosure yields statistically significant RankIC gains for effective-event-date $y_0$ and following-day $y_1$. Matched-versus-shuffled differences are also significant at both horizons, tying the gains to correct event-level context; analyst summaries provide no comparable improvement. Finance-specialized models show a group advantage on $y_0$ but no reliable advantage on $y_1$, while general-purpose Qwen models lead individually and scale effects are non-monotonic. LLM forecasts remain near consensus without consistently beating individual analysts; anchoring improves distributional alignment but worsens reported-EPS accuracy. Information-set construction, return horizon, model choice, and prediction interface therefore remain central to financial LLM evaluation.


\begin{table}[H]
\centering
\caption{Condensed prompt templates and output schemas for the three prediction interfaces.}
\label{tab:prompt-output-schema}

\begin{labelbox}{Shared Prompt Structure}{promptblue}
\footnotesize
\textbf{Role and task.} The model acts as a financial analyst and produces one prediction from the supplied information for the specified return-ranking or next-quarter EPS task.

\smallskip
\textbf{Conditional input blocks.} Only the blocks belonging to the evaluated information view are supplied:
\begin{itemize}[leftmargin=1.15em,itemsep=0pt,topsep=2pt,parsep=0pt]
    \item \texttt{[FUNDAMENTALS]}: cross-sectional ranks of current-quarter variables and year-over-year changes in revenue, earnings, margins, cash flow, leverage, and liquidity;
    \item \texttt{[ANALYST SUMMARY]}: pre-filing I/B/E/S target period, estimate count, EPS consensus mean and median, forecast dispersion, and recommendation statistics; and
    \item \texttt{[DISCLOSURE]}: the filed 10-Q MD\&A section.
\end{itemize}

\textbf{Evidence rule.} Use only the supplied blocks, return one final prediction, and tie the requested short evidence fields to those blocks.
\end{labelbox}

\vspace{-0.1em}
\begin{labelbox}{Return-Score Interface}{promptblue}
\footnotesize
\textbf{Task.} Produce a bounded score for expected short-horizon post-filing abnormal return; higher values indicate a more positive relative return signal.

\smallskip
\textbf{Decision rules.} Start from a neutral prior of zero and adjust once when the evidence justifies it. Positive and negative values indicate correspondingly positive and negative abnormal-return signals. Keep the score conservative unless evidence is strong; do not give alternative scenarios or revise the final score.

\smallskip
\textbf{Output (JSON only).}\par\noindent
\begin{minipage}{\linewidth}
\ttfamily\scriptsize
\{\newline
\hspace*{1em}"score": number in [-1.0, 1.0],\newline
\hspace*{1em}"confidence": number in [0.00, 1.00],\newline
\hspace*{1em}"evidence": ["short evidence 1", "short evidence 2"]\newline
\}
\end{minipage}
\end{labelbox}

\vspace{-0.1em}
\begin{labelbox}{Direct GPS Forecast Interface}{promptpink}
\footnotesize
\textbf{Task.} Forecast next-quarter GPS (GAAP/fully reported EPS) for the specified fiscal-period end, using the same scale as the I/B/E/S GPS forecasts.

\smallskip
\textbf{Decision rules.} Use cross-sectional fundamental ranks as operating context and consensus as a pre-filing expectation anchor. Focus on guidance, demand, pricing, margins, costs, liquidity, and risks affecting next-quarter earnings, and depart from consensus when the MD\&A supports a revision. Use the specified I/B/E/S units and return one numeric point forecast, not a trading score; the target date must match exactly.

\smallskip
\textbf{Required answer block.}\par\noindent
\begin{minipage}{\linewidth}
\ttfamily\scriptsize
\textless answer\textgreater\newline
MEASURE: GPS\newline
TARGET\_FPEDATS: YYYY-MM-DD\newline
FORECAST: float\newline
CONFIDENCE: float in [0.00, 1.00]\newline
EVIDENCE:\newline
- 2--4 concise evidence bullets\newline
\textless/answer\textgreater
\end{minipage}
\end{labelbox}

\vspace{-0.1em}
\begin{labelbox}{Consensus-Anchored GPS Interface}{promptpink}
\footnotesize
\textbf{Task.} Begin from the pre-filing analyst-mean GPS anchor and produce a single bounded adjustment. The evaluated point forecast is \texttt{anchor + delta}.

\smallskip
\textbf{Anchor fields.} The prompt supplies the target fiscal-period end and the analyst mean, median, and dispersion. The adjustment should remain close to zero unless the filing provides strong revision evidence.

\smallskip
\textbf{Decision rules.} Choose one delta around the analyst mean; when evidence is weak, keep it close to zero. Do not give alternative scenarios or output the final forecast directly, which is evaluated as \texttt{anchor + delta}.

\smallskip
\textbf{Output (JSON only).}\par\noindent
\begin{minipage}{\linewidth}
\ttfamily\scriptsize
\{\newline
\hspace*{1em}"delta": GPS EPS adjustment in [-3.0, 3.0],\newline
\hspace*{1em}"confidence": number in [0.00, 1.00],\newline
\hspace*{1em}"evidence": ["short evidence 1", "short evidence 2"]\newline
\}
\end{minipage}
\end{labelbox}

\end{table}

\clearpage
\balance
\bibliographystyle{ACM-Reference-Format}
\bibliography{custom,data_sources,longcontext}

\end{document}